\pdfoutput=1

\documentclass[11pt]{article}

\usepackage{emnlp2021}

\usepackage{times}
\usepackage{latexsym}
\usepackage{xcolor}

\usepackage{multirow}
\usepackage{graphicx}
\usepackage{stfloats}

\usepackage{amssymb}
\usepackage{amsmath}
\usepackage{tikz}
\usepackage[inline]{enumitem}

\usepackage[ruled,linesnumbered]{algorithm2e}

\usepackage{pifont}
\newcommand{\cmark}{\ding{51}}%
\newcommand{\xmark}{\ding{55}}%
\usepackage[T1]{fontenc}

\usepackage[utf8]{inputenc}

\usepackage{microtype}

\definecolor{colortext}{rgb}{1.0, 0.4, 0.0}
\definecolor{colortextdelete}{rgb}{0.0, 0.48, 0.45}

\definecolor{mossgreen}{rgb}{0.75, 0.75, 0.75}
\definecolor{sunset}{rgb}{0.5, 0.5, 0.5}
\definecolor{vividtangerine}{rgb}{0.25, 0.25, 0.25}

\newcommand{\GR}{\tikz\draw[mossgreen,fill=mossgreen] (0,0) circle (.6ex); }
\newcommand{\YL}{\tikz\draw[sunset,fill=sunset] (0,0) circle (.6ex); }
\newcommand{\RD}{\tikz\draw[vividtangerine,fill=vividtangerine] (0,0) circle (.6ex); }

\definecolor{mountainmeadow}{rgb}{0.19, 0.73, 0.56}
\definecolor{red(ncs)}{rgb}{0.77, 0.01, 0.2}

\title{From None to Severe: Predicting Severity in Movie Scripts}

\author{Yigeng Zhang\textsuperscript{\dag}, Mahsa Shafaei\textsuperscript{\dag}, Fabio A. González\textsuperscript{\ddag} \and Thamar Solorio\textsuperscript{\dag}\\
        \textsuperscript{\dag}University of Houston \\ \textsuperscript{\ddag}Universidad Nacional de Colombia\\
        \textsuperscript{\dag}\texttt{\{yzhang168,mshafaei,tsolorio\}@uh.edu} \\ \textsuperscript{\ddag}\texttt{fagonzalezo@unal.edu.co}}

\begin{document}
\maketitle
\begin{abstract}
In this paper, we introduce the task of predicting severity of age-restricted aspects of movie content based solely on the dialogue script. We first investigate categorizing the ordinal severity of movies on 5 aspects: \emph{Sex}, \emph{Violence}, \emph{Profanity}, \emph{Substance consumption}, and \emph{Frightening scenes}. The problem is handled using a Siamese network-based multitask framework which concurrently improves the interpretability of the predictions. The experimental results show that our method outperforms the previous state-of-the-art model and provides useful information to interpret model predictions. The proposed dataset and source code are publicly available at \href{https://github.com/RiTUAL-UH/Predicting-Severity-in-Movie-Scripts}{our GitHub repository}\footnote{\href{https://github.com/RiTUAL-UH/Predicting-Severity-in-Movie-Scripts}{https://github.com/RiTUAL-UH/Predicting-Severity-in-Movie-Scripts}.}.
\end{abstract}

\section{Introduction}
Estimating the severity level of objectionable content for movies can provide convenience for users to judge whether a movie is suitable for watching. For example, parents may want to make sure there are no violent scenes in a movie when they plan to watch it with their kids, because exposure to violent scenes may increase youth aggressive behavior and decrease their empathy \cite{anderson2017screen}.
However, existing rating systems (e.g., MPAA) only provide simple age restrictions and do not include the suitability level on a specific aspect of the content. 
Furthermore, a system that can automatically track the severity level of objectionable content helps the creative professionals evaluate the age suitability of their work. They may get assisted by this function and adjust the product creation or marketing strategy based on the corresponding target audiences. This system can be easily applied to any dialogue-intensive compositions like novel and screenplay writing. Content evaluation and intervention by the writers can happen at any stage of production to assess age-restricted contents.


In this work, we propose to solve the problem of predicting the severity of age-restricted content solely using the dialogue script data. 
Text is much more lightweight than visual data (such as images and videos), so the processing procedure can be more efficient and scalable considering the increasing fidelity of multimedia content. 
We initiate our exploration on movies from five aspects of contents: \emph{Sex \& Nudity}, \emph{Violence \& Gore}, \emph{Profanity}, \emph{Alcohol, Drugs \& Smoking}, and \emph{Frightening \& Intense Scenes} as used in \emph{IMDB}\footnote{\href{https://www.imdb.com/}{https://www.imdb.com/}, one of the most visited online databases of film, TV and celebrity content.} Parent Guide. 


There are a small number of previous works that studied modeling age-restricted content. \cite{shafaei-etal-2020-age} initiated the research of predicting MPAA ratings of the movies leveraging movie script and metadata. \cite{Martinez_Somandepalli_Singla_Ramakrishna_Uhls_Narayanan_2019} focused on violence detection using movie scripts while \cite{martinez-etal-2020-joint} expanded the scope to violence, substance abuse, and sex. 
Both works intended to predict the severity of age-restricted content into three manually defined levels: low, mid, and high. 
In this work, We introduce two more aspects of interest: \emph{Frightening} and \emph{Profanity}. Instead of manually downgrading severity levels into three categories, we explore with a more challenging setting: rating on 4 originally defined fine-grained severity levels from collective rating by customers: \emph{None}, \emph{Mild}, \emph{Moderate}, and \emph{Severe}. 

The major contributions of our research can be summarized as follows: 
\begin{enumerate*}[label={(\arabic*)}]
\item This work is the first attempt to solve the age-restricted content severity predicting problem from 5 aspects. We studied multiple baselines and presented a competitive method to inspire future exploration.
\item To our best knowledge, the dataset we developed is the first publicly available dataset for this task. The size is roughly five times larger than the restricted datasets from the previous works. 
\item We proposed an effective multitask ranking-classification framework to solve this problem. Our method dealt with long movie scripts successfully and achieved state-of-the-art results in all five aspects of age-restricted contents with rich interpretability.
\end{enumerate*}

\section{Methodology}
We investigate predicting the severity of 5 objectionable aspects of movies. This problem is formulated as a multi-class classification task. 
The average length of the dialogue scripts is around 10,000 words, which drastically exceeds the limit of current popular Transformer-based models. To leverage the strong semantic representation capability of the Transformers, we propose to represent each utterance as the basic unit, and further encode the context with the recurrent modules. Finally, we use a fully connected layer on top of the encoded representations to produce the classification predictions.

For this model, we first leverage SentenceTransformers \cite{reimers-gurevych-2019-sentence} to encode each dialogue utterance. Then, a Bi-directional LSTM encoder is deployed to model the sequential interrelations of the utterance flow. We finally apply a max-pooling operation on all time steps of the hidden states of the recurrent module to get the document representation for classification following the practice in \cite{howard-ruder-2018-universal}. We also study another strong word-level deep learning model, TextRCNN \cite{Lai_Xu_Liu_Zhao_2015}, to probe the significance of lexical signals.

\subsection{The Multitask Ranking-Classification Framework} 
The severity of a particular age-restricted content is a relative concept. People assign severity ratings to movies based on their own experiences and personal beliefs. Meanwhile, the severity levels are ordinal variables instead of independent categorical classes. Therefore, customers can gain a vivid understanding of the severity levels of an unfamiliar movie when comparing it to some examples (e.g., previous watched ones).

\begin{figure}[ht]
\centering
  \includegraphics[width=0.9\linewidth]{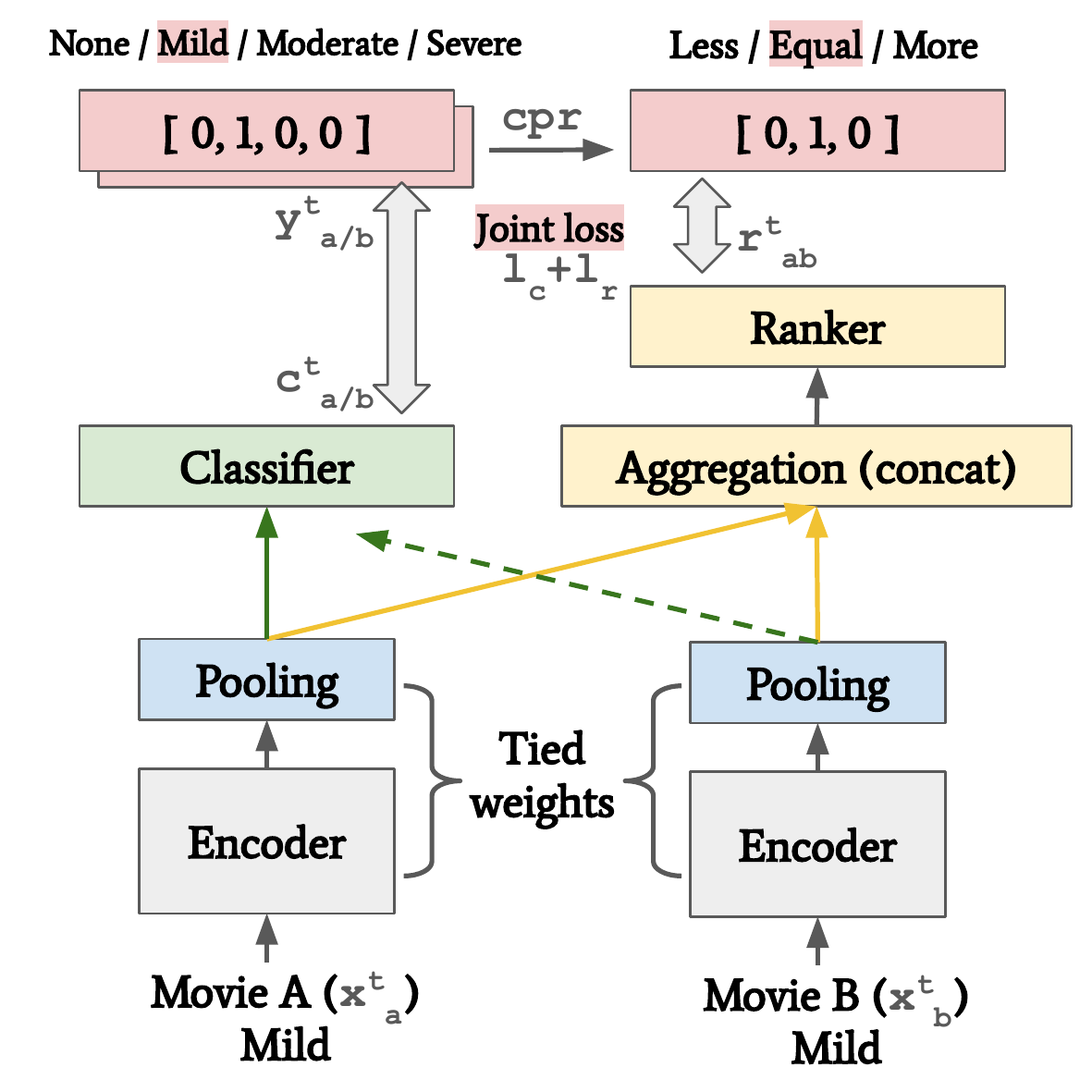}
  \caption{Multitask pairwise ranking-classification network.}
  \label{fig:siamese}
\end{figure}
The general model development algorithm is described as shown in Algorithm \ref{algorithm}.
We assume that learning to compare movies on their severity is a proxy to understand how the model differentiates severity. So we propose a pairwise ranking objective \cite{HULLERMEIER20081897} as an auxiliary task to probe into the model behavior for a more interpretable prediction. Other than existing multitask practices for text classification \cite{10.5555/3060832.3061023,ijcai2017-473}, this framework is based on a Siamese network with tied-weights for both instance classification and comparison, and it can be adopted by any backbone encoder. Here we apply it to the Bi-LSTM + Transformers model (RNN-Trans)  and TextRCNN model. The classification and the ranking objectives learn over two individual cross-entropy losses. Then the model is optimized on a joint loss of classification and ranking. The model structure is illustrated in Figure \ref{fig:siamese}.

\begin{table*}[]
\centering
\resizebox{0.85\textwidth}{!}{%
\begin{tabular}{c|l|ccccc|c}
\hline
\multicolumn{2}{c|}{Method}                                                                                         & Sex            & Violence       & Profanity      & Substance      & Frightening    & Avg            \\ \hline
\multirow{4}{*}{\begin{tabular}[c]{@{}c@{}}Baseline \\ model\end{tabular}}           & LR + Glove Avg.                 & 33.87          & 46.35          & 48.06          & 29.27          & 41.38          & 39.79          \\
                                                                                     & SVM + Glove Avg.                & 27.48          & 41.88          & 44.16          & 18.68          & 35.42          & 33.52          \\
                                                                                     & TextCNN                      & 38.19          & 46.61          & 63.95          & 37.16          & 44.82          & 46.14          \\
                                                                                    & BERT                      & 33.73          & 36.29          & 49.48          & 34.58          & 37.75          & 38.37          \\ \hline
\multirow{2}{*}{\begin{tabular}[c]{@{}c@{}}Backbone\\ model\end{tabular}}            & TextRCNN                     & 43.13          & 52.51          & 66.49          & 41.79          & 49.74          & 50.73          \\
                                                                                     & RNN-Trans             & 44.76          & 55.72          & 62.32          & 42.39          & 50.95          & 51.23          \\ \hline
\multirow{2}{*}{\begin{tabular}[c]{@{}c@{}}Proposed\\ multi-task model\end{tabular}} & TextRCNN S-MT        & 43.01          & 52.59          & \textbf{67.26} & \textbf{43.92} & 50.36          & 51.43          \\
                                                                                     & RNN-Trans S-MT & \textbf{45.68} & \textbf{55.90} & 62.65          & 42.21          & \textbf{51.55} & \textbf{51.60} \\ \hline
\end{tabular}%
}
\caption{Severity prediction macro F1 scores on test data.}
\label{tab:result}
\end{table*}

\begin{algorithm}[h]
\caption{Ranking-Classification}\label{algorithm}
\KwIn{training instance set $\mathbb{X}_{t}$ with severity label set $\mathbb{Y}_{t}$, ranking-classification model $f$, comparison operation $cpr$, classification/ranking loss $\mathcal{L}_{c}$/$\mathcal{L}_{r}$.}
\KwOut{multitask ranking-classification model   $\hat{f}$}
\SetKwFunction{FRank}{RANK-CLS}
  \SetKwProg{Pn}{Function}{:}{\KwRet}
  \Pn{\FRank{$f$, $\mathbb{X}_{t}$, $\mathbb{Y}_{t}$}}{
        initialization\;
        \While{not stopping criteria}{
        randomly pick $x^{t}_{i}$, $x^{t}_{j}\in \mathbb{X}_{t}$ with corresponding $y^{t}_{i}$, $y^{t}_{j}\in \mathbb{Y}_{t}$\;
        $c_{i, j}, r_{ij} \leftarrow f(x^{t}_{i}, x^{t}_{j})$\;
        $l_{c} \leftarrow \mathcal{L}_{c}(c_{i, j},y^{t}_{i, j})$\;
        $l_{r} \leftarrow \mathcal{L}_{r}(r_{ij}, cpr(y^{t}_{i},y^{t}_{j}))$\;
        $\hat{f} \leftarrow \underset{f}{argmin}$ $(l_{c} + l_{r}) $\;
        }
  }

\end{algorithm}

The auxiliary ranking component will distinguish the severity difference out of the training pairs, with a learning objective of predicting lower/equal/higher severity between the two instances. By introducing this function, we can apply model $f$ to compare the severity level of any pair of movies given an aspect.

\section{Dataset}
The dataset used in this work was developed based on the script data used in \cite{Shafaei2019,shafaei-etal-2020-age}. We collected the up-to-date user ratings for age-restricted content from \emph{IMDB.com} for more than 15,000 movies. The age-restricted aspects are adopted from the \emph{Parents Guide} section of each movies, and there are five aspects: \emph{Sex \& Nudity}, \emph{Violence \& Gore}, \emph{Profanity}, \emph{Alcohol, Drugs \& Smoking}, and \emph{Frightening \& Intense Scenes}. Each of the aspects has four severity levels for the users to rate on the corresponding movies from low to high, which are \emph{None}, \emph{Mild}, \emph{Moderate}, and \emph{Severe}. 
In this work, we pick the ratings on the website as the category label for each aspect. 

After collecting the user ratings, we filter out movies with less than 5 votes to make sure the collected severity level is robust to use. At last, we have roughly 4,400 to 6,600 movies for each aspect. 

The movie scripts have a median/average length of around 10,000 words. The vocabulary size of each aspect is roughly 330,000 to 450,000. Detailed dataset descriptions are attached in the appendix. 

Comparing to the previous works with restricted data access \cite{Martinez_Somandepalli_Singla_Ramakrishna_Uhls_Narayanan_2019, martinez-etal-2020-joint}, our dataset is roughly five times larger, and the data source is free to access. We made the updated dataset publicly available with the same data partitions in this work for reproducibility purposes.

\section{Experimental Results}

We evaluate the effectiveness of different methods on macro F1 score because the data is imbalanced.  The dataset is first split and stratified using an 80/10/10 ratio for training, development, and test for each age-restricted aspect. The experimental results on the test set are reported in Table \ref{tab:result}. Baseline models include average GloVe embedding \cite{pennington-etal-2014-glove} with SVM/Logistic Regression, TextCNN \cite{kim-2014-convolutional}, and BERT \cite{devlin-etal-2019-bert}. The proposed multitask model outperforms multiple baselines by a compelling margin in all aspects. Statistical significance test shows introducing the ranking subtask does no harm to model performance.

\begin{table}[]
\centering
\resizebox{\columnwidth}{!}{%
\begin{tabular}{l|c|c|c|c|c}
\hline
Method  & Sex.  & Vio.  & Pro.  & Sub.  & Fri.  \\ \hline
\cite{shafaei-etal-2020-age}      & 29.21      & 36.65      & 50.57      & 33.48 & 27.82 \\ \hline
\cite{martinez-etal-2020-joint}  & 40.91 & 53.02 & 60.51 & 35.60 & 48.81 \\ \hline
TextRCNN S-MT & 41.27 & 54.11 & \textbf{69.51} & \textbf{43.56} & 47.18 \\ \hline
RNN-Trans S-MT & \textbf{44.66} & \textbf{55.29} & 64.01 & 42.63 & \textbf{51.03} \\ \hline

\end{tabular}%
}
\caption{Performance benchmarking with related approaches over the same 10-fold cross-validation.}
\label{tab:compare}
\end{table}


\begin{table*}[]
\renewcommand\arraystretch{1.2}
\centering
\resizebox{\textwidth}{!}{%
\begin{tabular}{|l||l|l|l||l|l|l|l|}
\hline
Movie                       & Aspect      & Gold label & Prediction & None                                                                                       & Mild                                                                                       & Moderate                                                                                   & Severe                                                                                     \\ \hline\hline
Deadpool (2016)             & Profanity   & Severe     & Severe \cmark     & \RD\RD\RD\RD\RD & \RD\RD\RD\RD\RD & \RD\RD\RD\RD\RD & \RD\YL\YL\YL\GR \\ \hline
Pride \& Prejudice (2005)           & Violence    & None       & None \cmark       & \RD\YL\YL\YL\YL & \YL\YL\YL\GR\GR & \GR\GR\GR\GR\GR & \GR\GR\GR\GR\GR \\ \hline
Django Unchained (2012)     & Sex      & Moderate   & Mild \xmark       & \RD\RD\RD\RD\RD & \YL\YL\YL\YL\YL & \YL\GR\GR\GR\GR & \GR\GR\GR\GR\GR \\ \hline
A Clockwork Orange (1971)   & Substance     & Moderate   & Mild \xmark      & \RD\RD\RD\RD\RD & \YL\YL\YL\YL\YL & \GR\GR\GR\GR\GR & \GR\GR\GR\GR\GR \\ \hline
The Greatest Showman (2017) & Frightening & Mild       & Mild \cmark      & \RD\RD\RD\RD\RD & \YL\YL\GR\GR\GR & \GR\GR\GR\GR\GR & \GR\GR\GR\GR\GR \\ \hline
\end{tabular}%
}
\caption{Analysis on successful/unsuccessful test examples. Predictions with \cmark\:mean correct while with \xmark\:represents incorrect. For candidate comparators in each severity level, a black circle indicates the model believes the test sample has a higher severity level than the comparator. Similarly, a gray circle means the test sample has an equal severity level, and a 
light gray circle means lower severity. The comparison results come from best-performing models of each aspect.}
\label{tab:case-study}
\end{table*}

The proposed RNN-Trans multitask Siamese model  (RNN-Trans S-MT) dominates on \emph{Sex \& Nudity}, \emph{Violence \& Gore}, and \emph{Frightening \& intensive scenes} while the proposed TextRCNN multitask Siamese model (TextRCNN S-MT) works best in \emph{Profanity} and \emph{Alcohol, Drugs \& Smoking}. This outcome is intuitive because \emph{Profanity} is the only aspect that has an overt pattern such as bad words and abusive language included in the dialogue script. Those utterances are neither missing nor latent to both audiences and the NLP model. So word-level models can have advantages over the utterance-level model in catching such signals. As for \emph{Frightening \& intensive scenes} and \emph{Violence \& Gore}, they are more challenging for the model to make inference because the data lacks visual and audio information. Dialogues might sometimes imply scenes such as a violent fight with cursing words or a frightening shot with screaming, however, not all such scenes come with particular dialogues. For \emph{Sex \& Nudity} and \emph{Alcohol, Drugs \& Smoking}, it is even more difficult to infer from plain text without any visual evidence. 

We also experiment with models from related previous works by \cite{shafaei-etal-2020-age, martinez-etal-2020-joint}. The former leverages LSTM with attention and NRC emotion to model the textual signal to predict MPAA ratings for the movie with the script; the latter uses RNN on dialogue encoded by a pretrained MovieBERT model and sentiment embeddings to predict severity of different aspects simultaneously. For fair and reliable comparison, we conduct the benchmarking within the same setting of using the dialogue script as the only available information in the same 10-fold cross-validation configuration from \cite{martinez-etal-2020-joint}. Table \ref{tab:compare} shows our proposed method outperformed previous works by a large margin. 


To conclude, our proposed method works reliably better in \emph{Frightening \& intensive scenes} and \emph{Violence \& Gore} due to Transformer's strong ability to capture latent implications behind utterances. While the word-level model, TextRCNN, performs better in detecting overt signals in \emph{Profanity}. For \emph{Sex \& Nudity} and \emph{Alcohol, Drugs \& Smoking}, our method still achieved the state-of-the-art although modeling the severity for these aspects remains challenging.

\section{Discussion and Analysis}
We investigate several popular movies with at least 200,000 IMDB ratings from the test set of each aspect. We collect the top 5 movies with the largest number of severity ratings from each severity level as comparators, then do the pairwise ranking between each movie and all its comparators for test. Comparison results are shown in Table \ref{tab:case-study}.

\textbf{Successful examples:} For the movie \emph{Deadpool (2016)}, the model gives a correct prediction on the \emph{Profanity} aspect with a convincing signal from pairwise ranking. This movie is determined to have a higher severity level in profanity than any candidate comparators from \emph{None/Mild/Moderate} level. For \emph{Severe}-level comparators, the test instance is lower than one, equal to three, and higher than one of the comparators. The movie \emph{Pride \& Prejudice (2005)} comes with a correct prediction on \emph{Violence}. Four comparators from  \emph{None} show they have equal severity while only one denotes higher. Three \emph{Mild} comparators give equal results while two indicate the test movie is lower than them. For moderate and severe comparators, all of them evince the test movie has lower severity. We can therefore reason that \emph{Pride \& Prejudice (2005)} has a severity level of \emph{None} in terms of violence content. 
There is a similar case as presented in \emph{The Greatest Showman (2017)}. It tends to appear higher than none but roughly \emph{Mild} level among the comparators. Then the model gives it a correct prediction to \emph{Mild}. Therefore, by investigating the pairwise ranking results, the interpretation of this prediction is more vivid to make sense to users concerning the relative severity level.

\textbf{Unsuccessful examples:} For the movie \emph{Django Unchained (2012)} on \emph{Nudity} and the movie \emph{A Clockwork Orange (1971)} on \emph{Alcohol, Drugs \& Smoking}, the ranking gives a very clear pattern: higher than none, equal to mild, and lower than moderate. However, the actual severity level is moderate instead of mild. We conservatively argue aspects such as \emph{Nudity} and \emph{Alcohol, Drugs \& Smoking} can hardly be inferred from dialogue text. It tends to be unsurprising if there are adult scenes or drinking scenes existing in a movie without any verbal indication.

\section{Conclusion \& Future Work}
In this work, we applied a deep learning-based method to predict severity of age-restricted content based on movie script data. The experimental results show the proposed multi-task ranking-classification model outperforms the previous state-of-the-art method and can give rich interpretability by demonstrating severity using example comparator movies. Our work provides a reasonable groundbreaking exploration in this research topic for the community. For future work, we propose to investigate other modalities to capture relevant patterns and fine-grained aspects like violence types.

\section*{Acknowledgements}
This work was partially supported by the National Science Foundation under grant \# 2036368. We would like to thank the anonymous EMNLP reviewers for their feedback on this work.

\bibliography{anthology,custom}
\bibliographystyle{acl_natbib}

\clearpage
\newpage
\appendix
\section{Appendix}
\subsection{Document Length Distribution}
\begin{figure*}[b]
\centering
  \includegraphics[width=\linewidth]{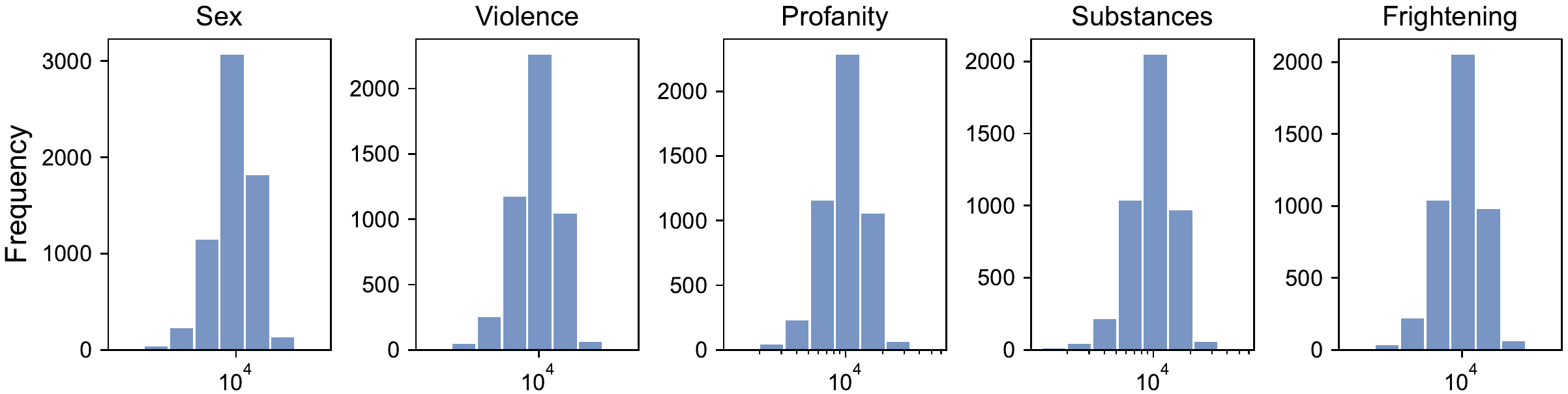}
  \caption{Document length distribution for each aspect. The plots use a logarithmic scale on the x-axis (document length).}
  \label{fig:len}
\end{figure*}

The document length distribution of each aspect is shown in Figure \ref{fig:len}. Different aspects share a similar document length distribution. The average length and the median length of the movie scripts are around 10,000 words.

\subsection{Label Distribution}
\begin{figure*}[b]
\centering
  \includegraphics[width=\linewidth]{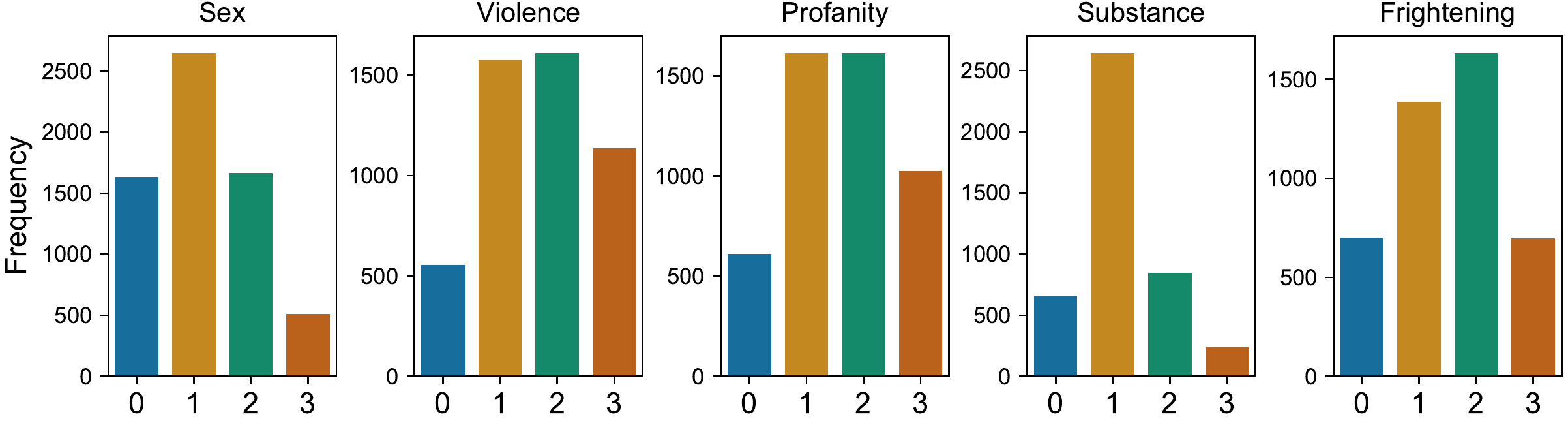}
  \caption{Label distribution for each aspect. 0-\emph{None}, 1-\emph{Mild}, 2-\emph{Moderate}, 3-\emph{Severe}.}
  \label{fig:label}
\end{figure*}

The severity label distribution for each aspect is unbalanced to a greater or lesser extent, as shown in Figure \ref{fig:label}.

\subsection{Data Separation}
The training, development, and test separation of each age-restricted aspect is shown in Table \ref{tab:number-aspect}.
\begin{table}[h]
\centering
\resizebox{\columnwidth}{!}{%
\begin{tabular}{l|ccccc}
\hline
Aspect & Sex & Violence & Profanity & Substance & Frightening \\ \hline
Train  & 5200   & 3921     & 3910      & 3538    & 3553        \\
Dev    & 651    & 491      & 489       & 443     & 445         \\
Test   & 651    & 491      & 489       & 443     & 445         \\ \hline
\end{tabular}%
}
\caption{The number of instances for each aspect.}
\label{tab:number-aspect}
\end{table}

\subsection{Computing Infrastructure \& Settings of Experiments}
We implement neural network models using PyTorch 1.6.0 and PyTorch Lightning 1.0.2. The experiments are executed on NVIDIA Tesla P40.

For model development, we use Glove embedding of 300-dimension trained on Wikipedia 2014 + Gigaword 5 (6B tokens) for TextCNN, TextRCNN models. Three 2D convolution modules of TextCNN are with kernel sizes 3, 4, 5 respectively. They all have 1 input channel and 10 output channels. BERT model uses the \emph{bert-base-uncased} model provided by the Python library \emph{Transformers}. The proposed model use sentence embeddings of 768-dimension from SentenceTransformer based on BERT-large. Bi-LSTM used in the proposed model and TextRCNN has a hidden size of 200 for each direction. All experiments with neural networks use Adam optimizer to optimize with a learning rate set to 0.001. 


\end{document}